\documentclass[10pt,twocolumn,letterpaper]{article}

\usepackage[pagenumbers]{wacv} 

\usepackage{graphicx}
\usepackage{amsmath}
\usepackage{amssymb}
\usepackage{booktabs}

\usepackage{amsmath}
\usepackage{caption}
\usepackage{subcaption}
\usepackage{multirow}
\usepackage[pagebackref,breaklinks,colorlinks]{hyperref}

\usepackage[capitalize]{cleveref}
\crefname{section}{Sec.}{Secs.}
\Crefname{section}{Section}{Sections}
\Crefname{table}{Table}{Tables}
\crefname{table}{Tab.}{Tabs.}

\def\wacvPaperID{2} 
\def\confName{WACV}
\def\confYear{2024}

\begin{document}

\title{How does Contrastive Learning Organize Images?}

\author{Yunzhe Zhang\\
Zhejiang University of Technology\\
288 Liuhe Rd\\
{\tt\small YunzheZh@outlook.com}
\and
Yao Lu\\
{\tt\small yaolu.zjut@gmail.com}
\and
Qi Xuan\\
{\tt\small xuanqi@zjut.edu.cn}
}
\maketitle

\begin{abstract}
Contrastive learning, a dominant self-supervised technique, emphasizes similarity in representations between augmentations of the same input and dissimilarity for different ones. Although low contrastive loss often correlates with high classification accuracy, recent studies challenge this direct relationship, spotlighting the crucial role of inductive biases. We delve into these biases from a clustering viewpoint, noting that contrastive learning creates locally dense clusters, contrasting the globally dense clusters from supervised learning. To capture this discrepancy, we introduce the "RLD (Relative Local Density)" metric. While this cluster property can hinder linear classification accuracy, leveraging a Graph Convolutional Network (GCN) based classifier mitigates this, boosting accuracy and reducing parameter requirements. The code is available \href{https://github.com/xsgxlz/How-does-Contrastive-Learning-Organize-Images/tree/main}{here}.
\end{abstract}

\section{Introduction}
In recent years, contrastive learning has revolutionized machine learning by effectively learning representation functions from unlabeled data. This method generates simple augmentations for each data point and enforces, through an appropriate loss function, that (1) augmentations of a single data point are clustered together and (2) augmentations of different data points remain distant. These representations yield competitive classification performance, even when using a linear classifier for various downstream tasks. This progress brings us closer to the long-standing aspiration of developing machine learning models capable of generalizing across diverse data distributions and tasks.

Researchers currently lack a conceptual framework for understanding such phenomena, which is not surprising considering the limited quantitative understanding of generalization for a single task and data distribution. However, even a partial conceptual understanding could advance the field, and researchers have started to address this challenge \cite{arora2019theoretical, tosh2021contrastive, haochen2021provable, wang2022chaos}.

Recent papers aim to guide the further development of this emerging theory by (1) analyzing the properties of augmented views' distributions and making assumptions, and (2) mathematically proving that, given these properties, an optimal contrastive loss can guarantee linear classification accuracy to some extent. Initial works in this area, such as \cite{saunshi2019theoretical} and \cite{lee2021predicting}, assume that two positive samples, as augmented views of the same image, are (nearly) conditionally independent given the class \(y\). In other words, augmented views from the same class share the same distribution, even if they are derived from different images. More recent works, like \cite{wang2022chaos}, adopt a more practical and weaker assumption: for images from the same class, their distributions of augmented views overlap, while those from different classes do not.

Papers following this paradigm exhibit a limitation. Although the mathematical proofs are rigorous, there is no guarantee that the assumptions made are correct. These assumptions are often based on intuition, which is understandable given our limited knowledge of the underlying data distributions. However, this means that such assumptions may not hold true in real-world scenarios. For example, \cite{saunshi2022understanding} points out that the assumptions in \cite{wang2022chaos} may not be valid, as there is little overlap between the distributions of augmented views in practice, regardless of the class they belong to.

The paper \cite{saunshi2022understanding} also demonstrates that when augmentation distributions for inputs do not overlap (which almost holds in practice), any discussion that does not consider inductive bias will lead to vacuous guarantees on classification accuracy. This raises a natural question: what does optimal contrastive loss mean in real-world scenarios when inductive bias is taken into account?

To address this question, we investigate the role of inductive bias in optimal contrastive loss from a clustering perspective and argue that a Graph Convolutional Network (GCN)-based classifier \cite{kipf2016semi} is more suitable for contrastive learning than a linear one. In summary, we make the following observations:

\begin{itemize}
\item Visually similar images have similar vector representations in contrastive learning, regardless of whether they belong to the same class. Conversely, images from the same class that are visually different have very distinct representations.
\item Due to this, all images from the same class form locally dense clusters in contrastive learning, as most images visually similar to a specific image usually belong to the same class. However, visually different images with the same class label remain distinct in vector space, resulting in clusters that are not globally dense. To quantitatively measure this property, we develop a metric called "RLD (Relative Local Density)" that assesses density in graph space.
\item Considering that contrastive learning creates dense clusters (communities) in graph space, we employ a GCN classifier, which not only achieves higher accuracy compared to the linear classifier but also requires fewer parameters.
\end{itemize}

\section{Related Works}
In this section, we discuss the underlying principles of contrastive learning and review relevant literature on this topic.

Contrastive learning has achieved remarkable success in solving downstream tasks by learning representations from similar pairs of data obtained using temporal information \cite{wang2015unsupervised, logeswaran2018efficient} or different views or augmentations of inputs \cite{hjelm2018learning, gao2021simcse, he2020momentum, dosovitskiy2014discriminative, chen2021exploring, wu2018unsupervised, tian2020contrastive, chen2020simple, bachman2019learning}. By minimizing a contrastive loss \cite{hadsell2006dimensionality, oord2018representation}, the similarity of a representation across various 'views' of the same image is maximized, while minimizing their similarity with distracting negative samples. 

Multiple views of a single data point can be naturally extracted from multimodal or multisensory data \cite{arandjelovic2017look, korbar2018cooperative, miech2019howto100m, owens2018audio, recasens2021broaden, sun2019videobert}, while for an image-only modality, they are typically constructed via local and global cropping \cite{bachman2019learning, henaff2020data, hjelm2018learning, oord2018representation} or data augmentation \cite{chen2020simple, doersch2017multi, dosovitskiy2014discriminative, he2020momentum, wu2018unsupervised}. Positive pairs correspond to views of the same data point, while negatives are sampled views of different data points (typically from the same mini-batch), although the necessity for negative samples has recently been questioned \cite{chen2021exploring, grill2020bootstrap, zbontar2021barlow}.

With the rapid development of contrastive learning, a number of studies have attempted to understand it. In line with the framework discussed in the introduction, \cite{saunshi2019theoretical} and \cite{lee2021predicting} initially assumed that positive samples are (nearly) conditionally independent, and then established a bound connecting contrastive loss and classification performance. In \cite{wang2022chaos}, the author posited an overlap between intra-class augmented views and provided a bound for downstream performance in relation to optimal contrastive loss.

However, \cite{saunshi2022understanding} contends that neglecting inductive biases of the function class and training algorithm fails to adequately explain the success of contrastive learning, and can even provably result in vacuous guarantees in certain situations. Furthermore, the study also asserts that there is minimal overlap between the distributions of augmented images, which contradicts basic assumptions made in numerous previous works.

With this understanding of contrastive learning and the existing literature, we will now proceed to explore the role of inductive biases and their impact on clustering and classification performance.

\section{Preliminaries}

\subsection{Cluster and Community Evaluation}
In this paper, we define clusters as distinct subsets within a data partition, where each subset represents a group of similar data points. Cluster evaluation has been a long-standing question in the literature \cite{hubert1985comparing, strehl2002cluster}, focusing on determining whether clusters are well-structured \cite{rousseeuw1987silhouettes} or consistent with ground truth labels \cite{rosenberg2007v}. In this study, we primarily concentrate on the former aspect, as ground truth labels themselves are treated as a partition, instead of those generated by the data-based cluster method.

The conventional notion of well-structured clusters, adopted in numerous works, is described as "dense and well-separated". A quantitative description of this concept is the Calinski-Harabasz score, also known as the Variance Ratio Criterion, introduced by \cite{calinski1974dendrite}. Given a set of data points \(\{x_1, x_2, ...,x_n \}\) and the cluster indices they belong to \(\{y_1, y_2, ...,y_n \}\), the Calinski-Harabasz score can be defined as:
\begin{equation}
\text{CH score} = \frac{\sum_{i=1}^{k} n_i \lVert \mu_i - \mu \rVert^2 / (k - 1)}{\sum_{i=1}^{k} \sum_{x \in C_i} \lVert x - \mu_i \rVert^2 / (n - k)}
\end{equation}
where \(k\) is the number of clusters, \(n\) is the total number of data points, \(n_i\) is the number of data points in cluster \(C_i\), \(\mu_i\) is the centroid of cluster \(C_i\), and \(\mu\) is the centroid of all data points. In essence, the Calinski-Harabasz score is the ratio of the sum of between-cluster dispersion to the sum of within-cluster dispersion for all clusters, where dispersion is defined as the sum of squared distances.

However, as we will illustrate later, the Calinski-Harabasz score, based on the conventional concept of clusters, does not correspond to the clusters formed by contrastive learning, necessitating the consideration of an alternative notion of clusters.

In this context, communities can be viewed as analogs to clusters but are based on graph space rather than Euclidean space, offering increased flexibility. The most commonly used metric for assessing community quality is modularity \cite{clauset2004finding}, which quantifies the density of connections within communities relative to the expected density of connections between randomly placed nodes. For graph \(G\) with adjacency matrix \(A \in \mathbb{R}^{N \times N}\), modularity \(Q\) is:
\begin{equation}
Q = \frac{1}{2m} \sum_{i,j} \left[ A_{ij} - \frac{k_i k_j}{2m} \right] \delta(y_i, y_j)
\label{eq:Q}
\end{equation}
where \(m\) is the sum of edge weights in the graph, \(k_i\) and \(k_j\) are the degrees of nodes \(i\) and \(j\), respectively, \(y_i\) and \(y_j\) are the community assignments of nodes \(i\) and \(j\), and \(\delta(y_i, y_j)\) is the Kronecker delta function, which equals 1 if \(y_i = y_j\) and 0 otherwise. Our experiments demonstrate that, with the proper construction of  \(G\), modularity can better reveal underlying cluster structures in contrastive learning.

\subsection{Graph Convolutional Network}
Graph Convolutional Networks (GCNs) are a class of neural networks specifically designed for handling graph-structured data. They have gained considerable attention in recent years due to their ability to learn and generalize information from graph data \cite{kipf2016semi, bruna2014spectral, defferrard2016convolutional}.

The term GCN can refer to all graph convolutional networks or a specific architecture of a graph convolutional network proposed by \cite{kipf2016semi}. In this paper, we focus on the latter for simplicity, as it serves as the origin of many other GCN models and captures the key characteristics of the broader class of GCN models.

Typical components of graph-structured data include node features \(X \in \mathbb{R}^{N \times C}\) and an adjacency matrix \(A \in \mathbb{R}^{N \times N}\), where \(N\) refers to the number of nodes and \(C\) refers to the dimensions of node features. The adjacency matrix captures the relationships between nodes in the graph.

A GCN aims to learn a function that maps the input node features \(X\) to a new set of node features \(Z \in \mathbb{R}^{N \times F}\), where \(F\) is the output feature dimension. The GCN proposed by \cite{kipf2016semi} performs graph convolutions in the following manner:
\begin{equation}
Z = \sigma \left(\tilde{A} X W \right),
\end{equation}
where \(\tilde{A} = \hat{D}^{-\frac{1}{2}} \hat{A} \hat{D}^{-\frac{1}{2}}\), with \(\hat{A} = A + I\) being the adjacency matrix with added self-connections and \(\hat{D}\) is the diagonal degree matrix of \(\hat{A}\), having elements \(\hat{D}_{ii} = \sum_j \hat{A}_{ij}\). The matrix \(W \in \mathbb{R}^{C \times F}\) is a learnable weight matrix, and \(\sigma (\cdot)\) is the activation function applied element-wise.

For deeper GCN architectures with \(L\) layers, the output feature matrix \(Z^{(L)}\) can be computed through a series of graph convolutions:
\begin{equation}
Z^{(l)} = \sigma \left(\tilde{A} Z^{(l-1)} W^{(l)} \right),
\end{equation}
where \(l = 1, \dots, L\), \(Z^{(0)} = X\), and \(W^{(l)} \in \mathbb{R}^{F_{l-1} \times F_l}\) are the learnable weight matrices for each layer. The dimensions of the input and output features change through layers, with \(F_0 = C\) and \(F_L = F\).

In this work, we introduce a minor modification to the GCN by incorporating a learnable scale parameter \(\alpha\) in each layer, which transforms the adjacency matrix as \(\hat{A} = \alpha A + I_N\). This adaptation enables the model to dynamically balance the significance of self-connections and connections to neighboring nodes, potentially enhancing the network's capability to extract valuable information from the graph structure. Consequently, the updated adjacency matrix \(\tilde{A}\) becomes:
\begin{equation}
\tilde{A} = \hat{D}^{-\frac{1}{2}} (\alpha A + I_N) \hat{D}^{-\frac{1}{2}},
\end{equation}
With this modified \(\tilde{A}\), the GCN can better adjust to the underlying structure of the graph during training, as it learns the optimal value of \(\alpha\) alongside the weight matrices \(W\).

We employ a GCN as the classifier due to its compatibility with graph space rather than Euclidean space, making it a better fit for the properties of clusters formed by contrastive learning. Experimental results demonstrate that the GCN consistently outperforms a linear classifier across various scenarios, further justifying its use in this context.

\begin{figure*}[t]
    \centering
    \begin{subfigure}[b]{0.45\textwidth}
        \centering
        \begin{subfigure}[b]{0.3\textwidth}
            \begin{subfigure}[b]{\textwidth}
                \includegraphics[width=\textwidth]{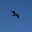}
            \end{subfigure}
            \begin{subfigure}[b]{\textwidth}
                \includegraphics[width=\textwidth]{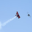}
            \end{subfigure}
        \caption{\\0.991\\0.566}
        \label{fig:1a}
        \end{subfigure}
        \begin{subfigure}[b]{0.3\textwidth}
            \begin{subfigure}[b]{\textwidth}
                \includegraphics[width=\textwidth]{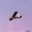}
            \end{subfigure}
            \begin{subfigure}[b]{\textwidth}
                \includegraphics[width=\textwidth]{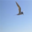}
            \end{subfigure}
        \caption{\\0.998\\0.276}
        \label{fig:1b}
        \end{subfigure}
        \begin{subfigure}[b]{0.3\textwidth}
            \begin{subfigure}[b]{\textwidth}
                \includegraphics[width=\textwidth]{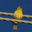}
            \end{subfigure}
            \begin{subfigure}[b]{\textwidth}
                \includegraphics[width=\textwidth]{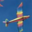}
            \end{subfigure}
        \caption{\\0.998\\-0.070}
        \label{fig:1c}
        \end{subfigure}
    \end{subfigure}
    \hfill
    \begin{minipage}[b]{0.09\textwidth}
        \centering
        \fontsize{8pt}{10pt}\selectfont
        Contrastive\\Supervised
        \vspace{0.07cm}
    \end{minipage}
    \hfill
    \begin{subfigure}[b]{0.45\textwidth}
        \centering
        \begin{subfigure}[b]{0.3\textwidth}
            \begin{subfigure}[b]{\textwidth}
                \includegraphics[width=\textwidth]{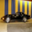}
            \end{subfigure}
            \begin{subfigure}[b]{\textwidth}
                \includegraphics[width=\textwidth]{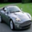}
            \end{subfigure}
        \caption{\\-0.745\\0.957}
        \label{fig:1d}
        \end{subfigure}
        \begin{subfigure}[b]{0.3\textwidth}
            \begin{subfigure}[b]{\textwidth}
                \includegraphics[width=\textwidth]{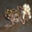}
            \end{subfigure}
            \begin{subfigure}[b]{\textwidth}
                \includegraphics[width=\textwidth]{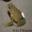}
            \end{subfigure}
        \caption{\\-0.722\\0.977}
        \label{fig:1e}
        \end{subfigure}
        \begin{subfigure}[b]{0.3\textwidth}
            \begin{subfigure}[b]{\textwidth}
                \includegraphics[width=\textwidth]{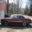}
            \end{subfigure}
            \begin{subfigure}[b]{\textwidth}
                \includegraphics[width=\textwidth]{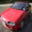}
            \end{subfigure}
        \caption{\\-0.554\\0.989}
        \label{fig:1f}
        \end{subfigure}
    \end{subfigure}
    \caption{\textbf{Exploring Image Pairs and Their Cosine Similarities in Representations.} \textbf{(a)} displays the image pair with the highest similarity across different labels in contrastive ResNet18, while \textbf{(b)} demonstrates the same for ResNet101, and \textbf{(c)} for ViT. In contrast, \textbf{(d)} illustrates the image pair with the lowest similarity within the same class in contrastive ResNet18, \textbf{(e)} for ResNet101, and \textbf{(f)} for ViT. The first line of the numbers beneath each image pair indicates their similarity in contrastive space, and the second line denotes the similarity in supervised space (using a supervised model with the same architecture). In \textbf{(a)}, the top image portrays a bird, and the bottom one depicts an airplane. The blue sky backgrounds and dark main bodies contribute to their visual similarity. Contrastive ResNet18 has difficulty distinguishing them, whereas supervised ResNet18 can. This observation also applies to \textbf{(b)} and \textbf{(c)}. Conversely, \textbf{(d)}, \textbf{(e)}, and \textbf{(f)} present pairs of visually distinct images (including hue, shape, perspective, etc.) within the same class, which are recognized by supervised models but not by contrastive models.}
    \label{fig:1}
\end{figure*}

\section{Experimental Settings}
In our analysis of contrastive learning models, several models were meticulously trained as subjects using standard procedures. This section delineates the experimental settings employed, leveraging the CIFAR-10 dataset \cite{krizhevsky2009learning}. We adopt the framework and augmentation proposed in \cite{chen2020improved} and incorporate the prediction head improvements suggested by \cite{dubois2022improving}, using a representation space dimension of 512. The model architectures tested include ResNet18, ResNet101, and a modified Vision Transformer with a parameter count similar to ResNet101 \cite{he2016deep, dosovitskiy2020image}. For comparison, we also train three models with identical architectures using supervised learning on CIFAR-10, maintaining a representation dimension of 512. All models presented in this paper utilize the AdamW optimizer \cite{loshchilov2017decoupled}. Further information regarding model architectures and hyperparameters can be found in the appendix. All experiments are done on one A100 GPU.

In subsequent sections, image vector representations are derived from raw images in the dataset using both contrastive and supervised models, consistent with standard evaluation methods. For contrastive models, all features are before the prediction head and normalized to unit vectors due to the undefined lengths during the training process.

We emphasize that our experimental settings are not exhaustively tuned for optimal performance. As such, achieving state-of-the-art results in \textbf{not our primary objective}. Instead, our main focus lies in uncovering novel insights into the inner workings of contrastive learning, rather than proposing practical improvements or techniques.

\section{Micro View: Image Pairs Under Contrastive Learning}
In this section, we delve into the distinct behaviors of contrastive and supervised learning methods in organizing images within the representation space. Our analysis encompasses a comprehensive examination of how these methods group images based on visual similarities and class labels. We begin by demonstrating how contrastive learning effectively groups visually similar images together, irrespective of their class labels. We then proceed to discuss the organization of images in supervised learning, which primarily focuses on class labels while often overlooking visual similarities within and across classes. Furthermore, we investigate the properties of k-nearest neighbors in the contrastive representation space and introduce the Class Homogeneity Index (CHI) to measure the average proportion of k-nearest neighbors that share the same class label. By comparing CHI across diverse models under contrastive and supervised learning, we provide insights into the local and global class-wise homogeneity of neighborhoods in the representation space and how architectural choices influence the behavior of contrastive learning.

\subsection{Contrastive Representation Space and Visual Similarity}
we analyze specific image pairs from the CIFAR-10 training dataset to illustrate the properties of contrastive representation space. \cref{fig:1a}, \cref{fig:1b}, and \cref{fig:1c} display image pairs with the highest cosine similarity in contrastive representation space, even though they belong to different classes. Conversely, \cref{fig:1d}, \cref{fig:1e}, and \cref{fig:1f} present image pairs with the lowest cosine similarity in contrastive representation space, despite having the same class label.

It is important to note that image pairs with high similarity in contrastive space exhibit visual resemblance, regardless of their class affiliation. On the other hand, image pairs that share the same label but display divergent visual appearances have relatively low similarity in contrastive space. This pattern is not observed in supervised space, where class labels rather than visual likenesses primarily influence similarity.

We hypothesize that this is a result of the unsupervised nature of contrastive learning, which relies solely on the information inherent in the samples. In contrast, supervised learning depends on the provided labels to organize images.

\begin{table*}[t]
\centering
\begin{tabular}{cllllll}
\toprule
\multirow{2}{*}{k} & \multicolumn{3}{c}{Contrastive} & \multicolumn{3}{c}{Supervised} \\ \cline{2-7} 
 &
  \multicolumn{1}{c}{ResNet18} &
  \multicolumn{1}{c}{ResNet101} &
  \multicolumn{1}{c}{ViT} &
  \multicolumn{1}{c}{ResNet18} &
  \multicolumn{1}{c}{ResNet101} &
  \multicolumn{1}{c}{ViT} \\ \cline{2-7} 
k=1                   & 0.8938    & 0.8818   & 0.8730   & 0.9680   & 0.9611   & 0.9903   \\
k=10                  & 0.8762    & 0.8662   & 0.8473   & 0.9674   & 0.9611   & 0.9893   \\
k=100                 & 0.8457    & 0.8438   & 0.8047   & 0.9669   & 0.9605   & 0.9886   \\
k=500                 & 0.7785    & 0.7945   & 0.6675   & 0.9666   & 0.9605   & 0.9881   \\
k=1000                & 0.7094    & 0.7267   & 0.5273   & 0.9663   & 0.9606   & 0.9879   \\
\bottomrule
\end{tabular}
\caption{\textbf{CHI across diverse models under contrastive and supervised learning.} The table presents the CHI for diverse models—ResNet18, ResNet101, and ViT—under both contrastive and supervised learning conditions, across different \(k)\) values. CHI measures the average proportion of k-nearest neighbors that share the same class label.}
\label{table:CHI}
\end{table*}

\subsection{K-Nearest Neighbors in Contrastive Representation Space}
As we have seen in the previous subsections, contrastive learning tends to group visually similar images together, regardless of their class labels. However, empirically, it is reasonable to assume that for most images, their visually similar counterparts belong to the same class. Building on our former observations and the insights from the analysis of visually similar image pairs, we can then infer that most representations have neighbors with the same label.

To show that, in this subsection, we delve into the properties of the k-nearest neighbors for each data point within the contrastive representation space. We use cosine similarity as our distance metric and compute the average proportion of k-nearest neighbors that share the same class, a metric we term the Class Homogeneity Index (CHI). We explore the effects of varying \(k\) values (\(k=1, 10, 100, 500, 1000\)) on CHI and summarize our findings in \cref{table:CHI}.

In light of the findings presented in \cref{table:CHI}, several implications emerge. First, for smaller \(k\) values, the high CHI observed in contrastive learning underscores that nearest neighbors typically belong to the same class. which exactly matches our earlier inference.

Second, the steep decrease in CHI as \(k\) increases in contrastive learning compared to supervised learning reveals an intriguing attribute. The neighborhood's class-wise homogeneity in the contrastive representation space seems to diminish more rapidly as we expand our neighborhood size. This implies that while contrastive features exhibit a local class-wise homogeneity, they do not extend this effect at a more global scale, unlike the supervised features.

Third, the most noticeable decline in CHI is observed for the ViT. This could indicate that its weaker inductive bias is a contributing factor. Therefore, the behavior of contrastive learning might be shaped not only by the data but also by architectural choices and their inherent inductive biases.

\section{Macro View: Contrastive Learning Forms Locally Dense Clusters}
Expanding from our previous analysis of individual images and their nearest neighbors within the representation space, we now widen our scope to a macroscopic perspective. The focus of this paper now shifts towards understanding how contrastive learning arranges data on a larger scale, specifically at the cluster level. Here, a "cluster" signifies the grouping formed by all data points belonging to the same class.

To characterize these clusters, we present a novel descriptor: "Locally Dense". A cluster is deemed "Locally Dense" if data points within the cluster are closely packed, creating a dense local neighborhood. Consequently, for a specific data point, its nearest neighbors are predominantly from its own cluster, leading to high class homogeneity within local regions of the cluster. To quantitatively measure the local density of a specific cluster, we propose a new graph-based metric called Relative Local Density (RLD).

Following this, we provide examples to illustrate what locally dense clusters appear like, and how they contrast from the typical "dense and well-separated" clusters. Importantly, "dense and well-separated" refers to clusters that are not just densely packed internally (locally dense), but are also clearly separated from other clusters, implying a global density and separation.

Lastly, we will substantiate that clusters formed by contrastive learning are locally dense but not globally, while clusters in supervised learning exhibit both local and global density and separation.

\begin{figure*}[t]
  \centering
  \includegraphics[width=0.9 \textwidth]{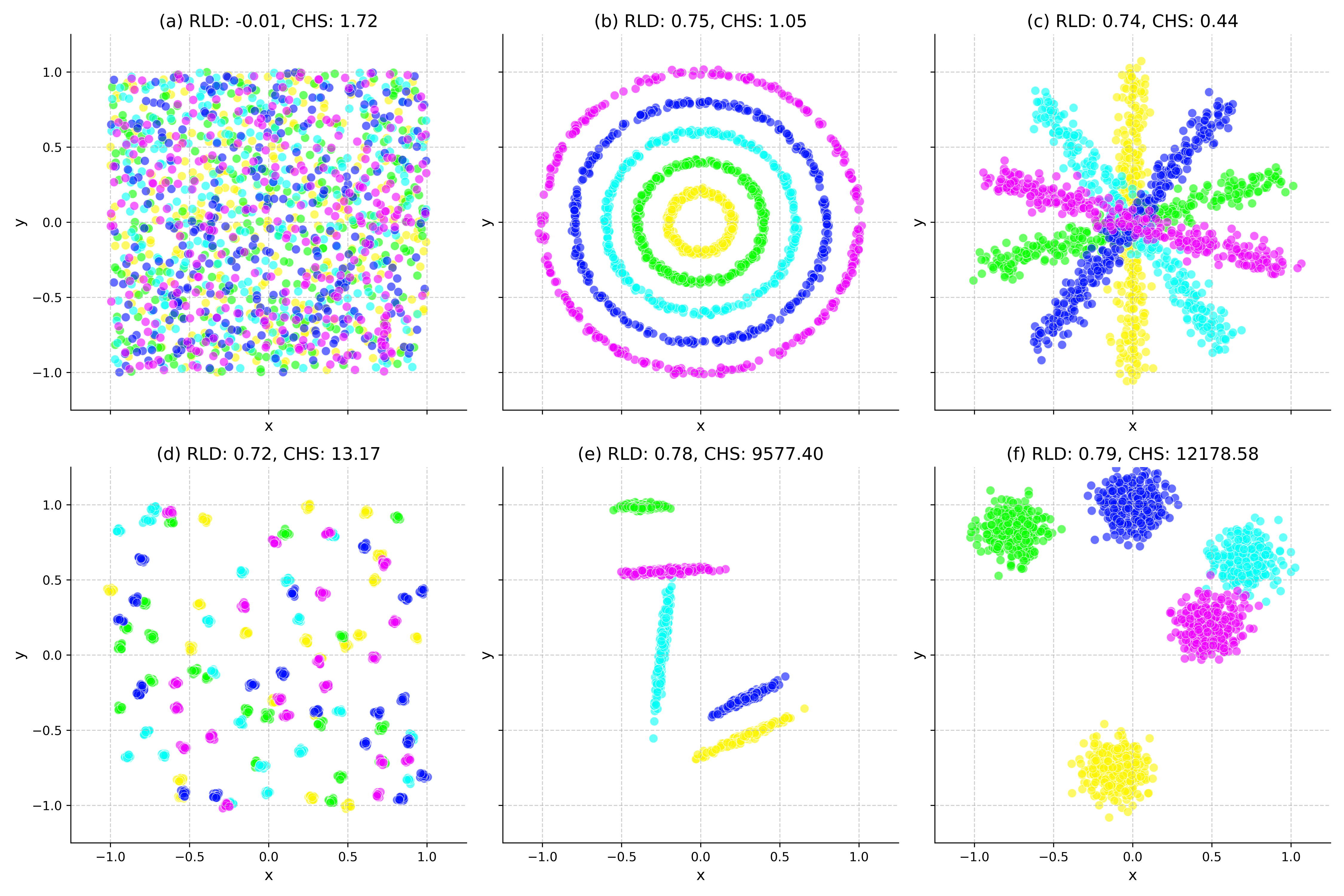}
  \caption{\textbf{Varied Cluster Configurations and Their Corresponding RLDs and CH scores.} Each subfigure, from (a) to (f), symbolizes a unique cluster configuration. Subfigure (a) presents data points uniformly scattered without discernible clustering, whereas (b) demonstrates a circular pattern with identifiable local clusters. Subfigure (c) showcases clusters arranged along distinct lines, and (d) displays small, tightly grouped local clusters dispersed throughout the plot. Subfigure (e) illustrates clusters formed along random lines, while (f) portrays multiple Gaussian distributions, each constituting a separate cluster. These examples underscore the varied characteristics of local and global densities and highlight the limitations of relying solely on global metrics, like the CH score, for evaluating cluster quality.}
  \label{fig:cluster examples}
\end{figure*}

\subsection{Relative Local Density: A Quantitative Measure for Locally Dense Clusters}
While the qualitative notion of locally dense clusters provides a conceptual foundation, it leaves room for a concrete, quantifiable measure. Thus, we introduce Relative Local Density (RLD), a novel metric aimed at capturing the notion of local density. This subsection details the construction of RLD, illuminating its correspondence to local density and its utility in comparing learning methods.

The computation of RLD involves several stages, each contributing to the encapsulation of local density. Initially, we construct a similarity matrix, \(S\), for a given set of data points \(X\). Each entry, \(S_{ij}\), captures the pairwise distance between data points \(X_i\) and \(X_j\), normalized by the mean distance and the square root of the dimension of \(X\) to ensure scale-invariance:

\begin{equation}
S_{ij} = \frac{-dist(X_{i}, X_{j})}{\sqrt{dim(X)} \cdot mean({dist(X_{p}, X_{q}): p \neq q})}
\end{equation}

This process converts distances into similarity scores, with higher scores indicating closer data points. The matrix diagonal elements are set to negative infinity, ensuring a data point does not regard itself as its neighbor.

The similarity matrix is then transformed into an adjacency matrix, \(A\), which encapsulates the relationships between data points in the feature space. A temperature parameter, \(T\), modulates this transformation, balancing the emphasis on local and global structures. A higher \(T\) yields a matrix with a more global structure, while a lower \(T\) retains more local information. The matrix is normalized, each entry divided by \(T\) and exponentially transformed to accentuate differences between data points:

\begin{equation}
A_{ij} = \frac{n^2 \cdot \exp(S_{ij}/T)^2}{\sum_{k=1}^{n} exp(S_{ik}/T)^2 \cdot \sum_{k=1}^{n} exp(S_{kj}/T)^2}
\end{equation}

\begin{figure*}[t]
\centering
\includegraphics[width=0.9 \textwidth]{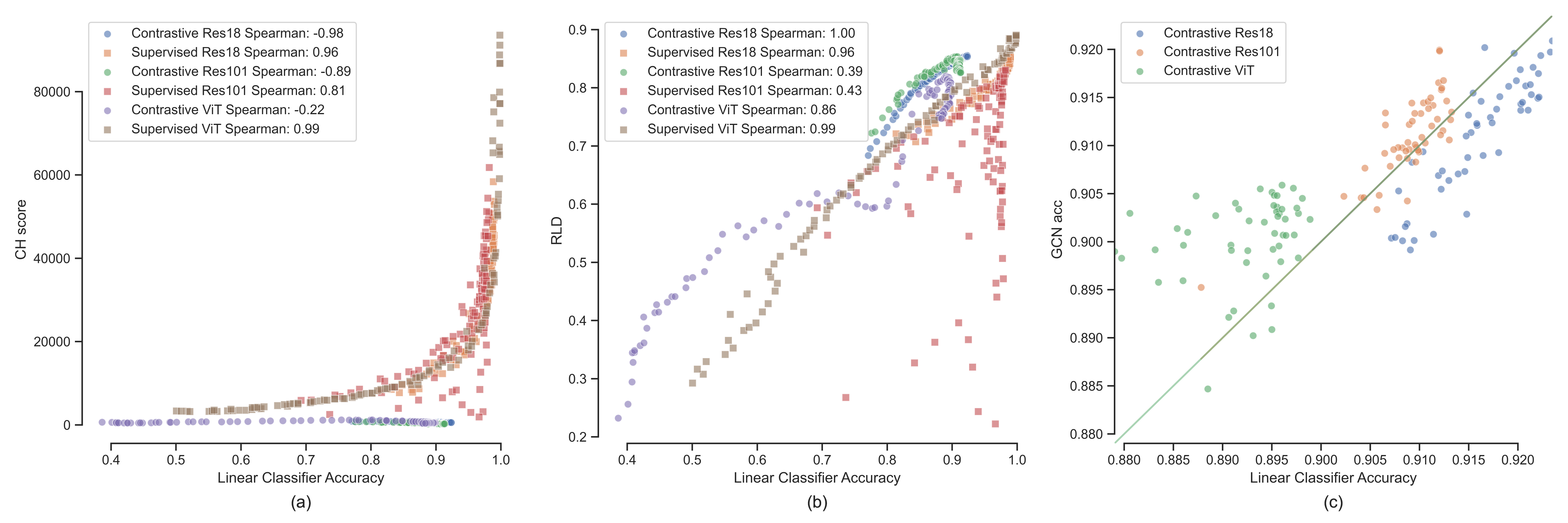}
\caption{\textbf{Comparative Analysis of Cluster Evaluation Metrics and Classifier Accuracy across Models.} The figure presents three scatter plots, each point represents a model, comparing different metrics with linear classifier accuracy for contrastive and supervised learning methods across three models: ResNet18, ResNet101, and ViT. (a) shows the correlation between CH scores and linear classifier accuracy. (b) illustrates the relationship between Relative Local Density (RLD) and linear classifier accuracy. (c) provides a comparison between GCN classifier accuracy and linear classifier accuracy for contrastive learning models.}
\label{fig:C and S}
\end{figure*}

The final RLD computation step measures the modularity of the adjacency matrix concerning class labels, \(y\), with \cref{eq:Q}. Modularity quantifies the strength of a graph's division into clusters. A high modularity score signifies dense intra-cluster connections and sparse inter-cluster connections, aligning with our local density intuition.

As a cluster evaluation metric, RLD offers several advantages:

1. \textbf{Local structure emphasis:} Unlike global metrics like the CH score, RLD captures local structure details, offering a more granular data organization understanding.
   
2. \textbf{Differentiability:} RLD's full differentiability enables integration into gradient-based optimization processes.
   
3. \textbf{Scale invariance:} RLD's insensitivity to the scale and number of data points enhances its versatility across diverse datasets.
   
4. \textbf{Graph techniques compatibility:} RLD's graph-based nature permits integration with other graph techniques, such as community detection.
   
5. \textbf{Comparability:} The modularity scaling to \((-0.5, 1)\) standardizes RLD, facilitating comparisons. Its application enables direct cluster comparison, as scores are relative to the same scale.

\subsection{Visualizing Locally Dense Clusters: Examples and Comparisons}
To gain a more intuitive understanding of local and global density concepts, it is important to visualize them before delving into the cluster analysis formed by contrastive and supervised learning methods. To facilitate this, we introduce illustrative examples that distinguish between locally dense clusters, as determined by RLD, and globally dense ones, as gauged by the CH score. 

\cref{fig:cluster examples} presents six distinct cluster examples, each illustrating the divergent nature of local and global densities. Let's examine some key takeaways:

1. A high RLD does not guarantee a high CH score. This fact is exemplified by clusters (b), (c), and (d), which, despite their high RLDs, register significantly lower CH scores compared to clusters (e) and (f).

2. The CH score fails to acknowledge the well-structured nature of clusters (b) and (c). This observation underscores the insensitivity of the CH score to certain types of cluster formations.

3. Clusters (b), (c), and (d) pose challenges for linear classifiers, which often struggle to define boundaries effectively. On the other hand, these classifiers are likely to perform well with clusters (e) and (f). Conversely, classifiers that operate in a neighbor-centric space, such as KNN, may deliver better separation results for clusters (b), (c), and (d).

\begin{figure*}[t]
\centering
\includegraphics[width=0.9 \textwidth]{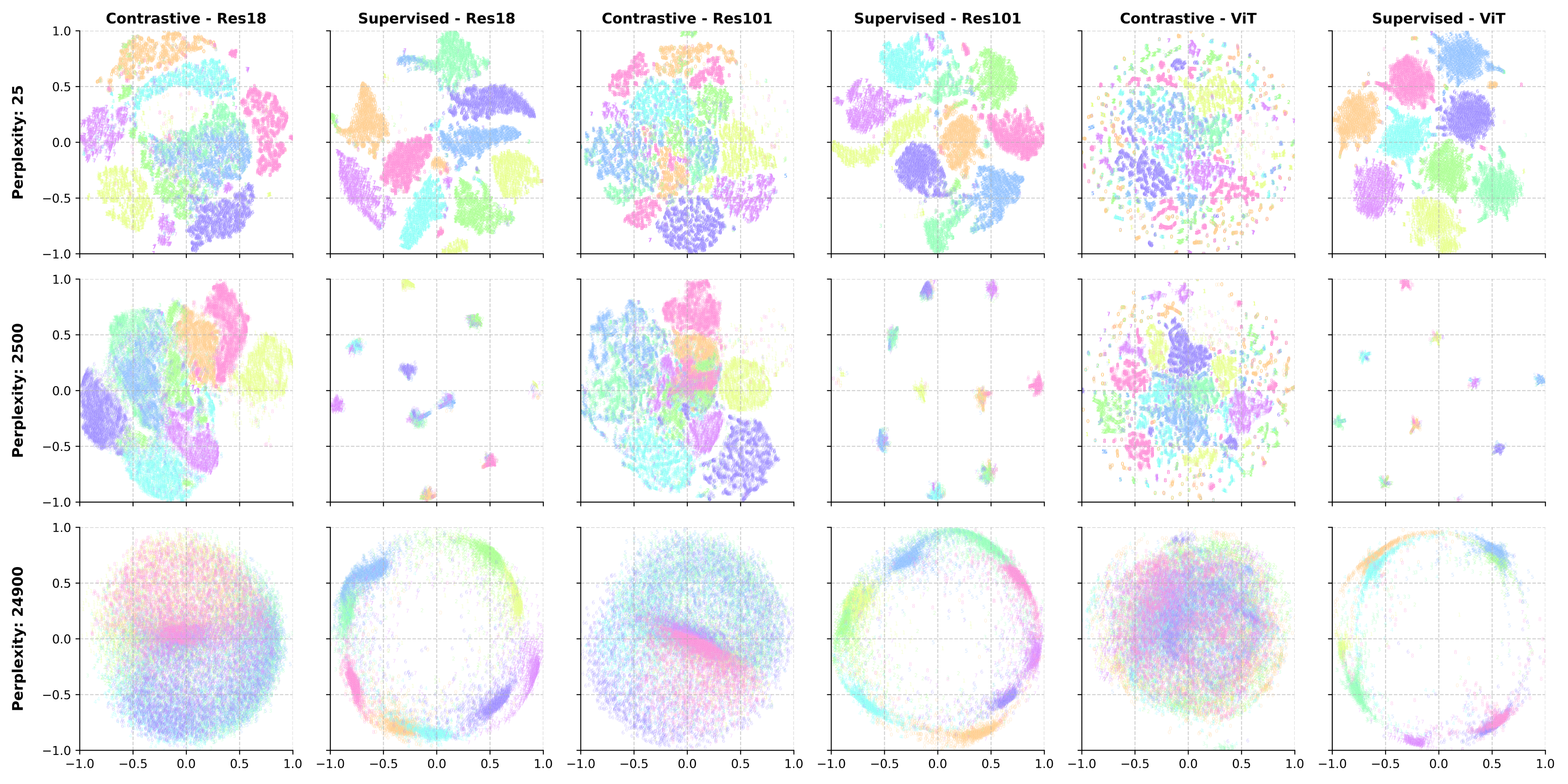}
\caption{\textbf{t-SNE Visualization of Normalized CIFAR-10 Features Generated by Contrastive and Supervised Learning.} This figure presents t-SNE visualizations of 25,000 data points for contrastive (left column) and supervised (right column) learning methods, across three different model architectures: ResNet18, ResNet101, and ViT. Each row corresponds to a different perplexity value (25, 2500, and 24900), showcasing the influence of this parameter on the visualization. As perplexity increases, contrastive clusters begin to dissipate, while supervised clusters remain relatively stable, highlighting the distinct clusters formed by the two learning methods.}
\label{fig:tsne}
\end{figure*}

\subsection{Comparing Clusters Formed by Contrastive Learning and Supervised Learning}
With a solid understanding of local and global density concepts, we can now turn our attention to examining the clusters produced by contrastive and supervised learning methods.

To provide a comprehensive and dynamic picture of cluster formation, we consider not only clusters created by fully trained models, but also those formed during the training process. This approach enables us to capture the evolution of clusters throughout the training phase and observe the unique ways in which different learning methods shape them over time.

In \cref{fig:C and S} (a) and (b), we present a real-case analysis. The RLDs (\(T=0.1\)) and CH scores for all contrastive and supervised features are displayed in relation to linear classifier accuracy, along with their respective Spearman correlation coefficients. This collective view provides a comparative perspective on how these learning methods impact data organization.

As demonstrated in \cref{fig:C and S} (a), there is a visible positive correlation between the CH score of supervised learning and linear classifier accuracy. In contrast, contrastive learning shows a negative correlation and maintains a nearly constant CH score throughout the entire training process. This observation suggests that contrastive learning does not create globally dense clusters.

However, as depicted in \cref{fig:C and S} (b), the Spearman correlation between linear classifier accuracy and RLD remains positive for both learning methods, given the same architecture. This result implies that both contrastive learning and supervised learning form locally dense clusters.

\paragraph{Applying a GCN Classifier} Given the challenges faced by linear classifiers in distinguishing locally dense clusters, as observed in \cref{fig:cluster examples} (b), (c), and (d), we decided to explore the use of a Graph Convolutional Network (GCN) classifier as an alternative. To facilitate this, we first constructed a graph identical to the one used in the computation of the RLDs.

When assigning node features, we utilized a one-hot encoding scheme. Each node was assigned a one-hot encoded vector of length equal to the number of classes, denoted as \(num\_classes\). In this scheme, each vector represented a specific class label, with the element corresponding to the class label set to 1, and all other elements set to 0.

Following the training method described in \cite{kipf2016semi}, we applied a mask rate of 0.2 and used a four-layer GCN. We then plotted the model accuracy in \cref{fig:C and S} (c), comparing it with linear accuracy. The results indicated that for the ResNet101 and ViT models, GCN accuracy was slightly higher than linear accuracy for most of the models, although the opposite was observed for the ResNet18 model. This suggests that a GCN classifier can sometimes outperform linear classifiers when dealing with clusters formed by contrastive learning methods.

\paragraph{Visual Evidence via t-SNE} Lastly, we employ t-SNE \cite{JMLR:v9:vandermaaten08a} as a visualization tool to further illustrate the differences between the features generated by contrastive and supervised learning methods. t-SNE is known for its ability to manage both 'local' and 'global' views via a tunable parameter known as perplexity. When we adjust this parameter in \cref{fig:tsne}, it becomes evident that contrastive clusters start to fade as the perplexity increases. This effect, however, is not observed in the clusters formed by supervised learning, thereby reinforcing our earlier observations about the distinct clusters formed by these two types of learning methods.

\section{Conclusion and Future Directions}
This study illuminates distinct differences in how contrastive and supervised learning algorithms structure data in representational space. We discover that contrastive learning primarily fosters 'locally dense' clusters while supervised learning generates globally dense clusters that align with class labels. The novel Relative Local Density (RLD) metric introduced in this study quantifies local density, offering a contrast to the traditional Calinski-Harabasz score. Implementing a Graph Convolutional Network (GCN) classifier demonstrates potential in tackling locally dense clusters, and the differences between learning methods are further highlighted through t-SNE visualizations.

Looking ahead, we propose two potential research directions. First, the development of more efficient classifiers tailored to clusters created by contrastive learning should be a priority. While GCNs show promise, their computational and memory requirements present challenges due to the need for an \(N \times N\) adjacency matrix and a \(N\) node feature matrix. Second, creating innovative augmentation algorithms could help prevent models from misclassifying visually similar images from different classes by distinguishing these images in augmented views.


{\small
\bibliographystyle{ieee_fullname}
\bibliography{egbib}
}

\end{document}